%% file: main.tex
\PassOptionsToPackage{table}{xcolor}
\PassOptionsToPackage{sort&compress}{natbib}
\PassOptionsToPackage{pagebackref,hypertexnames=false}{hyperref}
\documentclass{fairmeta}
\usepackage{amsmath,amssymb}
\usepackage{enumitem}
\usepackage{newtxtt}
\setcitestyle{numbers,square,comma}

\definecolor{oursbg}{RGB}{237,243,249}
\usepackage{pifont}
\newcommand{\yes}{\ding{51}} 
\newcommand{\no}{\ding{55}}  
\usepackage{adjustbox}
\usepackage{listings}
\usepackage{soul}
\definecolor{red1}{RGB}{255, 240, 240}
\definecolor{orange}{RGB}{231,151,68}
\definecolor{purple}{RGB}{158,108,205}
\definecolor{cyan}{RGB}{51,180,198}
\lstdefinestyle{scrollprompt}{
  basicstyle=\ttfamily\fontsize{6.5}{7.3}\selectfont,
  columns=fullflexible,
  breaklines=true,
  breakatwhitespace=false,
  keepspaces=true,
  frame=single,
  framerule=0.35pt,
  rulecolor=\color{black!38},
  backgroundcolor=\color{black!3},
  framesep=4pt,
  aboveskip=2pt,
  belowskip=2pt
}

\definecolor{scrollGain}{HTML}{B96B42}
\definecolor{scrollHeaderBg}{HTML}{EFF1F3}
\definecolor{scrollGroupBg}{HTML}{F5F6F7}
\definecolor{scrollStripeBg}{HTML}{FAFAFB}
\definecolor{scrollOursBg}{HTML}{F0F7FD}

\newcommand{\gain}[1]{%
  \nobreak\hspace{1pt}%
  \raisebox{-0.3ex}{%
    \textcolor{scrollGain}{\scriptsize$\uparrow\!#1$}%
  }%
}

\input{preamble}
\title{SCROLL: Lightweight Visual Indexing for On-Demand Evidence Exploration for Streaming Video Understanding}
\author[1,*]{Yaoyu Zhu}
\author[2,*]{Yudi Xie}
\author[3]{Zhengru Fang}
\author[3]{Jun Zhang}

\affiliation[1]{Southeast University}
\affiliation[2]{Wuhan University}
\affiliation[3]{The Hong Kong University of Science and Technology}
\contribution[*]{Equal contribution.}

\let\lababstract\abstract
\renewenvironment{abstract}{\noindent\textbf{Abstract}\par\smallskip\noindent\ignorespaces}{\unskip}
\lababstract{\input{sec/0_abstract}}
\newcommand{\maketitlesupplementary}{%
  \begin{center}
    \titlelist\par\medskip
    {\Large\sffamily Supplementary Material\par}
  \end{center}%
}

\begin{document}
\raggedbottom
\maketitle
\input{sec/1_intro}
\input{sec/2_related_work}

\input{sec/3_method}

\input{sec/4_experiment}

\input{sec/5_conclusion}
\clearpage
\input{sec/6_LLM_usage}\par
{
    \small
    \bibliographystyle{ieeenat_fullname}
    \bibliography{main}
}

\input{sec/X_suppl}

\end{document}

%% file: preamble.tex
\newcommand{\red}[1]{{\color{red}#1}}



%% file: sec/0_abstract.tex
\begin{abstract}
Streaming video understanding seeks to answer questions
about unfolding video streams, including events observed
long before a query arrives.
This requires continuously processing incoming frames
while keeping historical evidence accessible.
Existing approaches construct semantic memories through
repeated generative vision-language model (VLM) inference
or maintain visual memories through compression and selection.
The former incurs recurring streaming-time overhead,
while both may discard visual evidence needed for
unforeseen queries.
We argue that streaming memory should make historical
visual evidence efficiently accessible during streaming,
enabling exploration and interpretation on demand. To this end, we propose \textbf{SCROLL}, a training-free streaming video understanding framework that
separates lightweight streaming-time indexing from
adaptive query-time evidence exploration.
\textbf{SCROLL-Index} incrementally maintains a source-linked
visual index without generative VLM inference, while
\textbf{SCROLL-Explorer} uses an iterative Assess--Search--Keep
(ASK) process to assess evidence sufficiency, retrieve
historical observations, and selectively retain
visual context. On OVO-Bench with Qwen3-VL-8B, SCROLL achieves
74.00\% overall accuracy, averaged over real-time
perception and backward tracing, improving overall
and backward-tracing accuracy by 6.94 and 13.16
percentage points, respectively, over the Recent8
baseline. SCROLL also achieves 84.47\% on StreamingBench, while
SCROLL-Index maintains an average frame-ingestion latency
of around 0.25 seconds under continuous 1 FPS
input on a single RTX A5000 GPU. \red{The source code will be released.}
\end{abstract}

%% file: sec/1_intro.tex
\section{Introduction}
\label{sec:intro}
In streaming video understanding, systems continuously
observe an evolving video stream and answer questions
about ongoing and past events as they arise.
Unlike offline video understanding, they must process
incoming observations causally, without knowing which
events future queries will target.
Early online video systems have explored streaming
dialogue, persistent video perception, and active
interaction
\cite{chen2024videollmonline,huang2025ovbench,qian2025dispider}.
Yet causal processing alone does not guarantee timely
access to visual history: when memory updates fall
behind incoming frames, observed events may remain
unavailable when queried~\cite{ye2026venus}.
This makes maintaining historical access particularly
challenging when future queries are unknown during
ingestion~\cite{zhang2025flashvstream,minh2026protokv}.

\begin{figure}[t]
    \centering
    \includegraphics[width=0.7\linewidth]{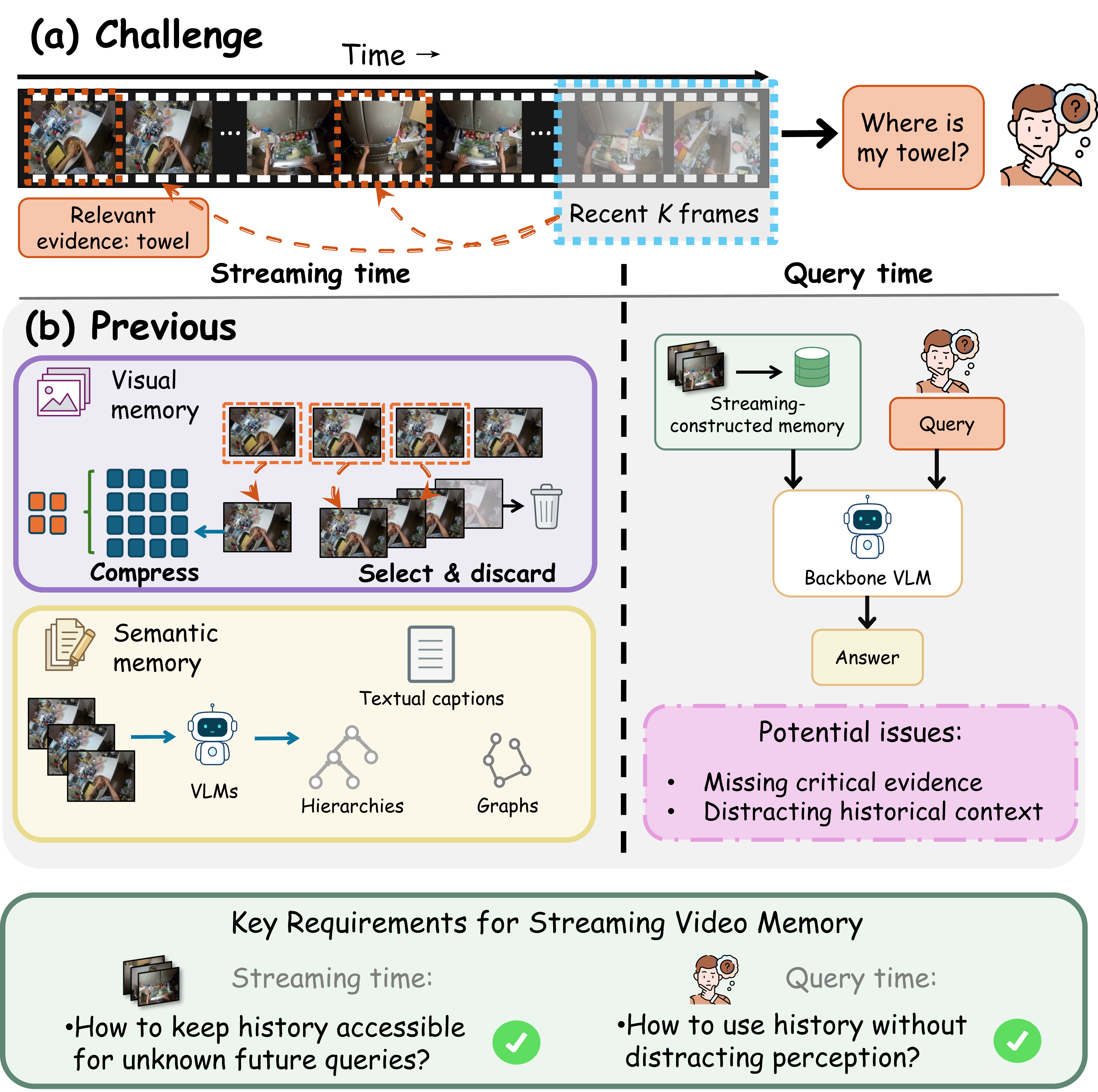}
    \caption{\textbf{Challenges and key requirements for streaming video memory.}
(a) Relevant evidence may appear long before a query arrives, beyond the recent observation window.
(b) Existing visual and semantic memory approaches process historical observations through compression, selection, or semantic abstraction, potentially introducing streaming-time overhead and losing useful details. Excessive historical context may also distract real-time perception.
Effective streaming memory requires timely access to historical visual evidence and selective evidence use at query time.}
    \label{fig:motivation}
    \vspace{-1.75em}
\end{figure}

Existing approaches maintain video history through
different memory strategies.
Semantic-memory methods construct textual observations,
event summaries, or hierarchies for retrieval
~\cite{oasis,dhsm}.
Such generative processing incurs recurring computational
overhead and may omit details needed for unforeseen queries.
Visual-memory methods compress or selectively maintain
visual representations
~\cite{xie2026fluxmem,savemem,selectstream}.
Although these strategies reduce stored visual context,
compression and selection can discard observations that
later become relevant.
When memory processing cannot keep pace with incoming
frames, it may further delay the availability of historical
evidence. Towards real-world streaming video understanding, the first question is: \sethlcolor{red1}
\textbf{\hl{How can streaming systems keep visual history accessible
when future queries are unknown?}}

\begin{figure}[t]
    \centering
    \includegraphics[width=0.65\linewidth]{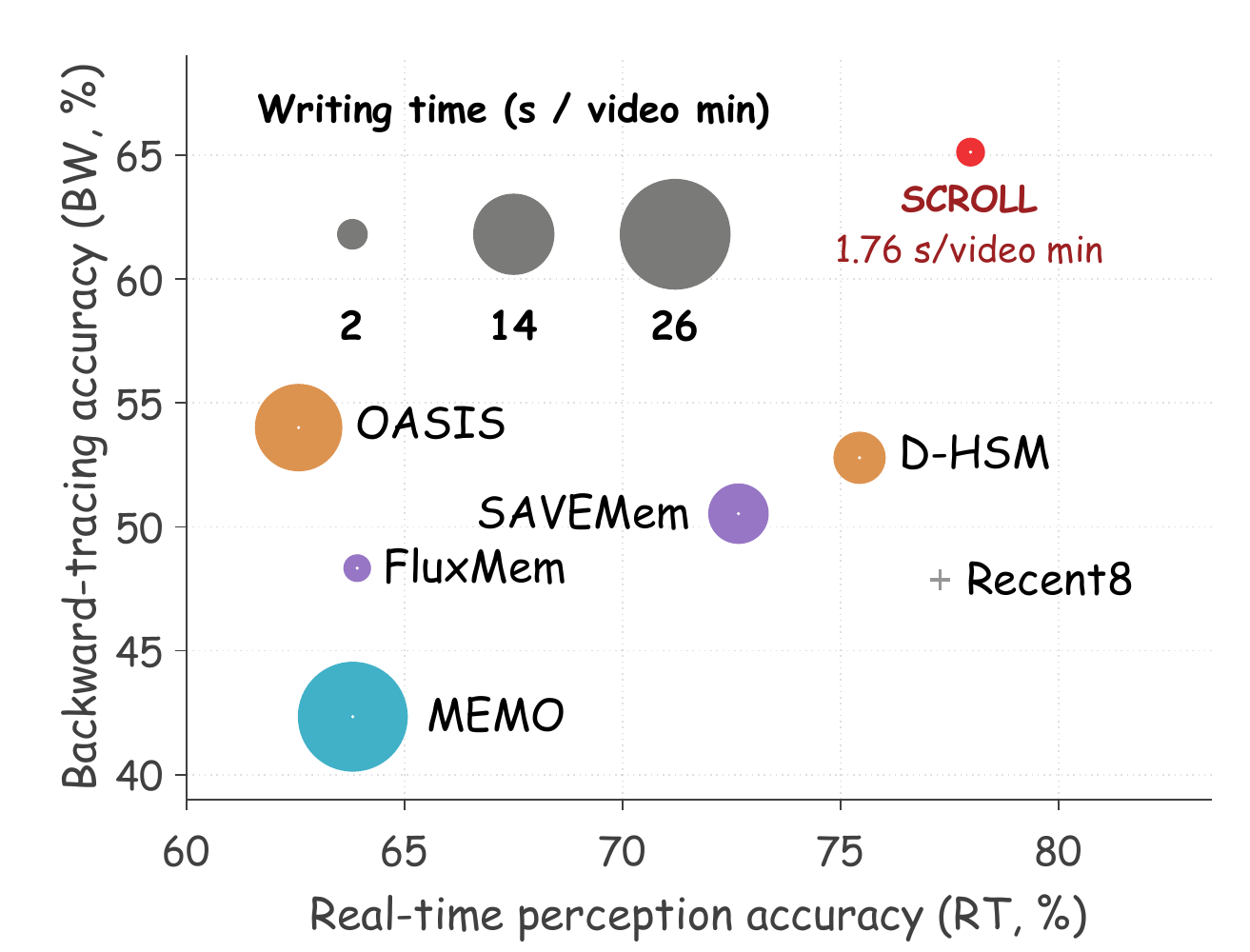}
\vspace{-1em}
\caption{\textbf{Accuracy--memory-writing-cost trade-off on OVO-Bench.}
The axes show real-time perception (RT) and backward-tracing
(BW) accuracy on Qwen-2.5-VL-7B.
Bubble area represents memory-writing time per video minute
(smaller is better).
SCROLL combines the highest BW and RT accuracy and low memory-writing cost amongst various baselines.}

    \label{fig:pareto}
    \vspace{-1.75em}
\end{figure}

Access to visual history alone, however, does not
guarantee better streaming understanding.
Recent studies show that simple recent-window baselines
can rival complex memory-elaborated systems, particularly in
real-time perception
~\cite{shen2026simplestream,selectstream}.
These methods maintain a focused view of the current
scene but cannot revisit history observations beyond their
recent window.
Conversely, additional historical context may improve
backward tracing while interfering with current-scene
understanding~\cite{shen2026simplestream}. This raises another question:
\sethlcolor{red1}
\textbf{\hl{How can streaming systems use historical evidence
without distracting real-time perception?}}

We investigate a complementary design in which
streaming memory prioritizes evidence accessibility
over exhaustive pre-query interpretation.
A lightweight visual index maintains searchable embeddings
and timestamps linked to original frames, keeping historical
observations accessible as the stream evolves without
requiring detailed semantic interpretation in advance.
At query time, the system assesses whether the available
visual context is sufficient, retrieves relevant historical
evidence when needed, and selectively retains useful
observations for subsequent reasoning.
Much like scrolling back through an unfolding record of
events, this design couples lightweight memory maintenance
with adaptive, evidence-guided exploration.

To this end, we propose \textbf{SCROLL}, a training-free
framework that decouples lightweight streaming-time indexing
from adaptive query-time evidence exploration.
\textbf{SCROLL-Index} incrementally maintains a source-linked
visual index without generative VLM inference, enabling
timely access to historical observations.
\textbf{SCROLL-Explorer} employs an iterative
Assess--Search--Keep (ASK) process that integrates
evidence sufficiency assessment, evidence-guided historical
search, and selective visual context retention.
Together, these components enable adaptive exploration of
historical evidence alongside recent visual context.
Figure~\ref{fig:pareto} compares SCROLL with generative
semantic memories (\textcolor{orange}{orange}), compressed
visual memories (\textcolor{purple}{purple}), and
structured visual indexing (\textcolor{cyan}{cyan}).
Recent8, which maintains recent-8 frames as memories, is shown
as a gray cross.
SCROLL (red) achieves the highest backward-tracing
accuracy among the compared methods while maintaining
strong real-time perception and low memory-writing cost.
Our contributions are as follows:



\begin{itemize}[
    leftmargin=2em,
    labelwidth=1.2em,
    labelsep=0.5em,
    itemsep=0.35em,
    topsep=0.3em,
    parsep=0pt
]

    \item[\raisebox{-0.2ex}{\includegraphics[width=1.15em]{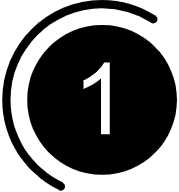}}]
    We introduce \textbf{SCROLL}, a training-free framework that separates continuous memory maintenance from query-conditioned interpretation. Its lightweight \textbf{SCROLL-Index} preserves timely, searchable access to source-linked visual evidence without generative memory construction.

    \item[\raisebox{-0.2ex}{\includegraphics[width=1.15em]{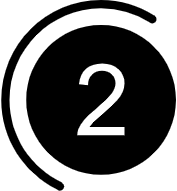}}]
    We develop \textbf{SCROLL-Explorer}, which integrates
    evidence sufficiency assessment, targeted historical
    retrieval, and selective visual context retention through
    an iterative ASK process, enabling adaptive evidence
    exploration without modifying the underlying VLM.

    \item[\raisebox{-0.2ex}{\includegraphics[width=1.15em]{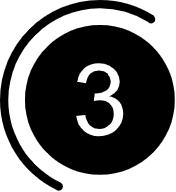}}]
    We evaluate SCROLL on OVO-Bench and StreamingBench
    across VLM backbones.
    Comprehensive ablation and efficiency experiments
    demonstrate substantial gains in historical understanding
    over recent-only inference, alongside strong real-time
    perception and low streaming-time memory construction cost.

\end{itemize}

%% file: sec/2_related_work.tex
\section{Related Work}
\label{sec:rw}

\begin{figure*}[t]
    \centering
    \includegraphics[width=\linewidth]{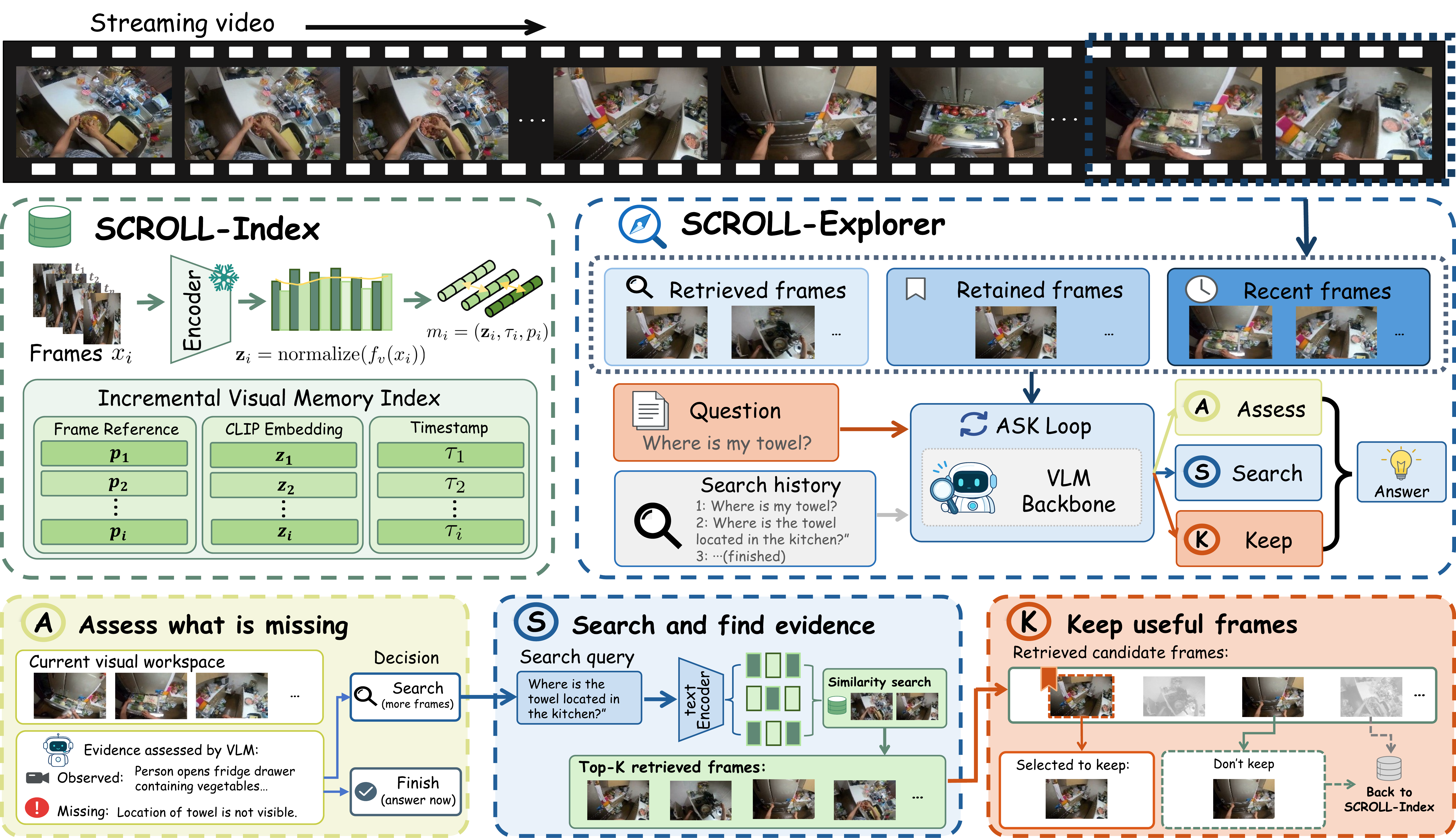}
    \caption{\textbf{Overview of the SCROLL framework.} During streaming, SCROLL-Index incrementally encodes sampled frames into lightweight visual embeddings, maintaining a searchable index linked to the original frames and timestamps without generative VLM inference. At query time, SCROLL-Explorer combines recent, retrieved, and retained frames within an iterative Assess–Search–Keep (ASK) loop. \textbf{Assess} evaluates evidence sufficiency and identifies missing information; \textbf{Search} retrieves relevant historical frames through query-conditioned similarity search; and \textbf{Keep} selectively retains useful evidence across exploration rounds. This design enables adaptive historical evidence acquisition within a bounded visual context while preserving access to the full indexed history.}
    \label{fig:placeholder}
    \vspace{-1.5em}
\end{figure*}

\subsection{Offline and Streaming Video Understanding}

Recent advances in long-video understanding are reflected in benchmarks such as Video-MME, LongVideoBench, and MLVU~\cite{fu2025videomme,wu2024longvideobench,zhou2025mlvu}.
Streaming video understanding further requires causal processing and timely responses to questions about ongoing and past events.
VideoLLM-online, VideoChat-Online, and Dispider explore online dialogue, perception, memory, and interaction~\cite{chen2024videollmonline,huang2025ovbench,qian2025dispider}.
OVO-Bench and StreamingBench evaluate these capabilities through temporally grounded questions~\cite{niu2025ovobench,lin2026streamingbench}.

\subsection{Streaming Video Memory Construction}

Prior video-memory systems maintain compact histories through
memory-augmented representations
\cite{he2024malmm,song2024moviechat,xiong2025streamchat,wang2025videollamb}.
Streaming methods also construct event-level or structured
semantic memories~\cite{zeng2025streamforest,dhsm,streamrag}.
Visual-memory approaches reduce computational costs through
KV-cache retrieval and compression
\cite{di2025rekv,kim2025infinipotv},
adaptive visual processing and temporal memory
\cite{shen2025longvu,zhang2026weavetime},
or selective memory maintenance
\cite{yang2025streammem,xie2026fluxmem,savemem,selectstream}.
Retrieval-oriented methods such as Vista and MEMO preserve
access to stored visual observations~\cite{vista,memo},
while ShallowStream constructs retrieval indices from
shallow MLLM representations and employs
query-time routing and evidence selection
for deeper inference~\cite{hao2026shallowstream}.
SCROLL instead maintains source-linked visual
embeddings and supports iterative,
evidence-conditioned exploration, where
the Reader can reassess information needs,
retrieve additional observations, and
selectively retain useful evidence across rounds.

\subsection{Agentic Video Evidence Retrieval}

Iterative retrieval interleaves reasoning and evidence
acquisition~\cite{trivedi2023ircot,asai2024selfrag}.
VideoAgent and DrVideo gather visual evidence over multiple
rounds~\cite{wang2024videoagent,drvideo},
while VideoTree and VCA explore adaptive video search
\cite{wang2025videotree,yang2025vca}.
VideoARM performs agentic reasoning over hierarchical
memory~\cite{yin2026videoarm}.
MERIT employs multi-key episodic retrieval~\cite{merit},
and AVP~\cite{avp} explores iterative evidence seeking
for long-video understanding.
For streaming video, OASIS retrieves from hierarchical
event memory~\cite{oasis}, StreamScout adapts the depth of
visual access for streaming queries~\cite{zhang2026streamscout},
and DeltaStreamer performs targeted historical replay
\cite{sun2026deltastreamer}.
SCROLL couples iterative evidence assessment and targeted
historical retrieval with selective visual retention,
operating over a continuously maintained index.
The ASK process manages recent, retrieved, and retained
frames within a bounded query-time visual context.

%% file: sec/3_method.tex
\section{Method}
\label{sec:method}

In streaming video understanding, visual observations arrive sequentially, and questions may be posed at arbitrary moments. Let $\mathcal{V}_t = \{(x_i,\tau_i)\mid\tau_i \leq t\}$ denote the observed video history at time $t$, where $x_i$ and $\tau_i$ represent a sampled frame and its timestamp, respectively. Given a question $q_t$, the system must produce an answer using only observations available up to time $t$. SCROLL separates continuous visual indexing from query-time evidence exploration. During streaming, a query-agnostic indexing procedure $F$ incrementally maintains a visual index: $\mathcal{I}_t = \mathcal{F}(\mathcal{V}_t)$, where each index entry links a visual representation to its stored source frame. At query time, an adaptive solver $G$ uses the recent frames $\mathcal{R}_t$ and the index $\mathcal{I}_t$ to retrieve historical visual evidence and produce an answer:
\begin{equation}
    \hat{a}_t = \mathcal{G}(q_t, R_t, I_t)
\end{equation}

\subsection{Lightweight Source-Linked Visual Indexing}
\label{sec:index}

SCROLL-Index processes incoming sampled frames using
a pretrained CLIP visual encoder~\cite{radford2021clip} $f_v$. For each frame $x_i$, the encoder extracts an $\ell_2$-normalized global visual embedding $\mathbf{z}_i \in \mathbb{R}^{d}$:
\begin{equation}
    \mathbf{z}_i =
    \frac{f_v(x_i)}{\|f_v(x_i)\|_2}.
\end{equation}
Each frame is encoded independently, and the index is updated as new embeddings become available. The streaming stage performs visual feature extraction and maintains references to the stored frames, without invoking a generative VLM to construct semantic descriptions.

Each embedding is associated with its timestamp $\tau_i$ and a source reference $p_i$ pointing to the corresponding stored frame. The index is incrementally updated as
\begin{equation}
    \mathcal{I} \leftarrow \mathcal{I} \cup
    \{(\mathbf{z}_i,\tau_i,p_i)\}.
\end{equation}
An entry becomes searchable once its encoding and index update are completed. The global embeddings serve as compact retrieval keys, while the source references provide access to the corresponding visual observations. The VLM backbone can subsequently examine selected historical frames using their stored visual content.

\subsection{Adaptive Query-Time Evidence Exploration}
\label{sec:explorer}

SCROLL-Explorer combines a VLM backbone with a frame-searching tool for historical retrieval and visual context management,
following the general paradigm of iterative tool-assisted
reasoning~\cite{yao2023react,wang2024videoagent,drvideo}.
Given a question $q_t$, it performs iterative
\emph{Assess--Search--Keep} (ASK) exploration using evidence collected across rounds to guide
subsequent decisions.

At round $r$, the visual context consists of three types of frames: \emph{Recent frames} $\mathcal{R}_t$ from the current observation window, \emph{Retrieved frames} $\mathcal{F}_r$ returned by the preceding search, and \emph{Retained frames} $\mathcal{H}_r$ preserved from earlier rounds. These collections are represented as lists of frames identified by unique IDs, with $\mathcal{H}_r$ maintaining the priority order used for retention. The visual context is assembled as
\begin{equation}
    \mathcal{C}_r =
    \operatorname{sort}_{\tau}
    \left(
        \operatorname{unique}
        (\mathcal{R}_t \Vert \mathcal{F}_r \Vert \mathcal{H}_r)
    \right),
    \label{eq:visual_context}
\end{equation}
where $\Vert$ denotes list concatenation, $\operatorname{unique}$ removes duplicate frame IDs, and $\operatorname{sort}_{\tau}$ orders the frames chronologically. Initially at round one, $\mathcal{F}_1=\mathcal{H}_1=\varnothing$, so the visual context contains only recent frames.

At each round, the VLM backbone receives the question, current visual context, and previous interaction history:
\begin{equation}
    \mathcal{D}_r =
    \mathrm{VLM}_{\theta}
    (q_t,\mathcal{C}_r,\mathcal{T}_{r-1}),
    \label{eq:vlm_decision}
\end{equation}
where $\mathcal{T}_{r-1}$ contains previous search requests, tool responses, and relevant evidence assessments. The structured output $\mathcal{D}_r$ includes visual observations $\mathcal{O}_r$, an information need $n_r$, a current answer, and an action $\alpha_r \in \{\texttt{search},\texttt{finish}\}$. It also specifies historical frame IDs selected for retention and, when searching, a search query with optional temporal constraints. These outputs are generated jointly in a single VLM backbone invocation. The tool executes the requested operations and updates the visual context for the next round.

\noindent \textbf{Evidence Sufficiency Assessment.}
At each round, the VLM backbone examines $\mathcal{C}_r$ to assess the evidence available for answering $q_t$. The assessment produces two complementary outputs: a set of observations $\mathcal{O}_r$ and an information need $n_r$ to guide subsequent search steps.

The observations describe relevant visual information in the current context, with references to the corresponding frames. They may identify objects, actions, scene attributes, or temporal relationships relevant to the question. The information need describes what remains unresolved, such as an object's earlier location, a missing action, or the ordering of two events. Relevant observations and information needs are carried forward through the interaction history to support subsequent decisions.

Based on the available evidence and previous interactions, the VLM backbone selects either \texttt{finish} or \texttt{search}. It can finish when the current context supports an answer, or continue exploring the history for additional evidence. The information need guides the next search query by identifying the specific information to investigate. The action is determined jointly with the evidence assessment and current answer, allowing the exploration process to adapt to the information available at each round.

\input{tables/ovo_comparison}
\input{tables/streaming_comparison}
\noindent \textbf{Evidence-Guided Historical Search.}
When the VLM backbone selects \texttt{search}, it generates a textual query $s_r$ conditioned on the original question, current observations, information need, and previous interaction history. The query can be updated across rounds as new visual evidence becomes available. The backbone may also specify a temporal anchor and a search direction (\texttt{before}, \texttt{after} or \texttt{both}) to focus retrieval on a relevant period.

The tool determines the eligible historical frames from $\mathcal{I}_t$. Let $\mathcal{B}_t$ denote the frame IDs preceding the recent observation window, and let $\mathcal{S}_r$ denote historical frame IDs previously retrieved or displayed during the current exploration. The temporal constraints define an eligible time range $\mathcal{W}_r$, bounded by the query time $t$. The candidate pool is
\begin{equation}
    \mathcal{P}_r =
    \left\{
        i \in \mathcal{B}_t
        \;\middle|\;
        i \notin \mathcal{S}_r,\;
        \tau_i \in \mathcal{W}_r
    \right\}.
    \label{eq:candidate_pool}
\end{equation}
The temporal range can cover the full eligible history or a restricted interval determined by the specified anchor and direction. The query-time cutoff remains fixed throughout the exploration of each question.

The tool encodes $s_r$ using the CLIP text encoder $f_{\mathrm{text}}$, producing a normalized query embedding $\mathbf{u}_r$. Candidate frames are ranked by their image-text similarity:
\begin{equation}
    \begin{aligned}
        \mathbf{u}_r &=
        \frac{f_{\mathrm{text}}(s_r)}
        {\|f_{\mathrm{text}}(s_r)\|_2},\\
        \rho_i^{(r)}
        &= \mathbf{u}_r^\top \mathbf{z}_i,
        \quad i \in \mathcal{P}_r.
    \end{aligned}
    \label{eq:retrieval_similarity}
\end{equation}
Here $\rho_i^{(r)}$ denotes the cosine similarity between the generated query and an indexed visual embedding. The tool ranks the eligible frames and selects a small number of candidates for visual examination.

To reduce visual redundancy, candidate selection also considers temporal diversity. The tool prioritizes high-scoring frames from distinct fixed-duration temporal bins, then fills any remaining retrieval slots with the next eligible candidates in the ranking. The selected frames are loaded using their stored source references $p_i$.

These frames form $\mathcal{F}_{r+1}$ and are presented to the VLM backbone in the next round. Their IDs are added to the visited-frame set $\mathcal{S}_{r+1}$, excluding them from subsequent searches. The VLM backbone can then examine the retrieved visual evidence and update its observations, information needs, and search queries in subsequent rounds.

\noindent \textbf{Selective Visual Context Retention.}
Historical frames obtained during exploration may remain relevant to subsequent decisions. SCROLL maintains a bounded set of Retained frames, allowing previously examined evidence to remain available as new frames are retrieved.

During each VLM invocation, the model selects historical frame IDs from the currently displayed Retrieved and Retained frames. The tool maps these to the corresponding frames, forming an ordered list $\mathcal{U}_r$. It updates the retained context by prioritizing the selected frames and filling the remaining capacity with previously retained frames:
\begin{equation}
    \mathcal{H}_{r+1} =
    \left[
        \operatorname{unique}
        (\mathcal{U}_r \Vert \mathcal{H}_r)
    \right]_{1:B},
    \label{eq:retention}
\end{equation}
where $[\cdot]_{1:B}$ denotes the first $B$ entries, or all entries if fewer than $B$ frames are available. The resulting list preserves the priority of the current selection while allowing earlier evidence to remain in the context when capacity permits.

Retention operates on historical frames already visible in $\mathcal{C}_r$. Frames returned by the current search enter $\mathcal{F}_{r+1}$ and become available for assessment and retention in the following round. Recent frames remain available independently of the historical retention budget $B$. Frames removed from the retained context remain stored in the visual index, while the visited-frame set prevents repeated retrieval during the same exploration.

When exploration continues, the tool assembles $\mathcal{C}_{r+1}$ from Recent, Retrieved, and Retained frames according to Eq.~\ref{eq:visual_context}. The process terminates when the VLM backbone selects \texttt{finish}, the maximum number of rounds is reached, or repeated searches return no new evidence. The answer is produced using the visual evidence available at termination.

%% file: tables/ovo_comparison.tex
\begin{table*}[t]
\centering
\caption{\textbf{Results on the real-time perception and backward tracing subsets of OVO-Bench.}
Accuracy (\%). RT and BW average respective tasks;
Overall is computed as $(\mathrm{RT}+\mathrm{BW})/2$.
$\dagger$ denotes baselines reproduced by us.
Bold and underline indicate best and runner-up scores
within each major category.}
\label{tab:01-ovo-main}

\begingroup
\fontsize{8.5}{10}\selectfont
\renewcommand{\arraystretch}{1.12}
\setlength{\tabcolsep}{2.75pt}

\begin{adjustbox}{max width=\textwidth}
\begin{tabular}{lcrrrrrrrrrrrr}
\hline
\rowcolor{scrollHeaderBg}
& &
\multicolumn{7}{c}{\textbf{Real-Time Visual Perception}}
& \multicolumn{4}{c}{\textbf{Backward Tracing}} & \\
\cline{3-9}\cline{10-13}
\rowcolor{scrollHeaderBg}
\multirow{-2}{*}{\textbf{Model}}
& \multirow{-2}{*}{\textbf{FPS}}
& \textbf{OCR} & \textbf{ACR} & \textbf{ATR}
& \textbf{STU} & \textbf{FPD} & \textbf{OJR}
& \textbf{RT} & \textbf{EPM} & \textbf{ASI}
& \textbf{HLD} & \textbf{BW}
& \multirow{-2}{*}{\textbf{Overall}} \\
\hline

Human~\cite{niu2025ovobench}
& -- & 93.96 & 92.57 & 94.83 & 92.70
& 91.09 & 94.02 & 93.20 & 92.59
& 93.02 & 91.37 & 92.33 & 92.77 \\

\hline
\rowcolor{scrollGroupBg}
\multicolumn{14}{l}{\textit{\textbf{Proprietary Models}}} \\

Gemini 1.5 Pro~\cite{oasis}
& 1
& \textbf{85.91} & \textbf{66.97}
& \textbf{79.31} & \textbf{58.43}
& \underline{63.37} & \textbf{61.96}
& \textbf{69.33} & \textbf{58.59}
& \textbf{76.35} & \textbf{52.64}
& \textbf{62.53} & \textbf{65.93} \\

\rowcolor{scrollStripeBg}
GPT-4o~\cite{huang2025ovbench}
& --
& \underline{69.80} & \underline{64.22}
& \underline{71.55} & \underline{51.12}
& \textbf{70.30} & \underline{59.78}
& \underline{64.46} & \underline{57.91}
& \underline{75.68} & \underline{48.66}
& \underline{60.75} & \underline{62.61} \\

\hline
\rowcolor{scrollGroupBg}
\multicolumn{14}{l}{
\textit{\textbf{Open-source Online MLLMs (Training-Based)}}} \\

SelectStream (8B)~\cite{selectstream}
& 1
& \textbf{94.60} & \textbf{86.70}
& \textbf{84.10} & \underline{68.36}
& \textbf{78.90} & \textbf{83.90}
& \textbf{82.76} & \textbf{62.40}
& 66.80 & \underline{57.40}
& \textbf{62.20} & \textbf{72.48} \\

\rowcolor{scrollStripeBg}
AURA (8B)~\cite{aura2026}
& 2
& \underline{89.9} & \underline{79.8}
& \underline{80.2} & \textbf{70.2}
& \underline{77.2} & \underline{81.5}
& \underline{79.8} & 54.9
& \underline{67.6} & \textbf{58.6}
& \underline{60.4} & \underline{70.08} \\

EventMemAgent (8B)~\cite{wen2026eventmemagent}
& 1
& 75.84 & 69.72 & 73.28 & 55.62
& 67.33 & 67.93 & 68.29
& \underline{59.60} & \textbf{70.95}
& 43.55 & 58.03 & 63.16 \\

\rowcolor{scrollStripeBg}
StreamForest (7B)~\cite{zeng2025streamforest}
& 1
& 68.46 & 53.21 & 71.55 & 47.75
& 65.35 & 60.87 & 61.20
& 58.92 & 64.86 & 32.26
& 52.01 & 56.61 \\

\hline
\rowcolor{scrollGroupBg}
\multicolumn{14}{l}{
\textit{\textbf{Open-source Online MLLMs (Training-Free)}}} \\

StreamScout (8B)~\cite{zhang2026streamscout}
& --
& 91.28 & 82.57 & 82.76 & 64.04
& 75.25 & 75.00 & 78.48
& \underline{55.22} & 61.49 & 63.44
& 60.05 & 69.27 \\

\rowcolor{scrollStripeBg}
SimpleStream (8B)~\cite{shen2026simplestream}
& 1
& \textbf{94.00} & \textbf{85.30}
& \underline{82.80} & 65.70
& \underline{77.20} & \textbf{83.20}
& \textbf{81.37} & 51.90 & 58.10
& 52.10 & 54.03 & 67.70 \\

HERMES (7B)~\cite{zhang2026hermes}
& 1
& 85.2 & 64.2 & 71.6 & 53.4
& 74.3 & 65.2 & 69.0
& 48.5 & 62.2 & 37.6
& 49.4 & 59.20 \\

\rowcolor{scrollGroupBg}
\multicolumn{14}{l}{
\textit{\textbf{Qwen2.5-VL-7B$^{\dagger}$}}~
\cite{bai2025qwen25vl}} \\

\hspace{0.9em}+Recent8
& 1
& 91.95 & 72.48 & 81.90 & 65.17
& 72.28 & 79.89 & 77.28
& \underline{55.22} & 58.78 & 29.57
& 47.86 & 62.57 \\

\rowcolor{scrollStripeBg}
\hspace{0.9em}+D-HSM~\cite{dhsm}
& 1
& 92.62 & 77.06 & 75.86 & 57.87
& 75.25 & 73.91 & 75.43
& 49.83 & 60.14 & 48.39
& 52.78 & 64.11 \\

\hspace{0.9em}+OASIS~\cite{oasis}
& 2
& 81.21 & 64.22 & 55.17 & 50.56
& 63.37 & 60.87 & 62.57
& 46.80 & 53.38 & 61.83
& 54.00 & 58.28 \\

\rowcolor{scrollStripeBg}
\hspace{0.9em}+FluxMem~\cite{xie2026fluxmem}
& 1
& 79.19 & 56.88 & 67.24 & 47.19
& 74.26 & 58.70 & 63.91
& 50.84 & 63.51 & 30.65
& 48.33 & 56.12 \\

\hspace{0.9em}+SAVEMem~\cite{savemem}
& 1
& 90.60 & 70.64 & 75.00 & 59.55
& 73.27 & 66.85 & 72.65
& 50.17 & 64.86 & 36.56
& 50.53 & 61.59 \\

\rowcolor{scrollOursBg}
\hspace{0.9em}\textbf{+SCROLL (Ours)}
& 1
& 92.62 & 77.98 & 81.90 & 60.11
& 74.26 & 80.98 & 77.97
& \textbf{57.91} & 62.16
& \textbf{75.27} & \underline{65.11}
& \underline{71.54} \\

\rowcolor{scrollGroupBg}
\multicolumn{14}{l}{
\textit{\textbf{Qwen3-VL-8B$^{\dagger}$}}~
\cite{bai2025qwen3vl}} \\

\hspace{0.9em}+Recent8
& 1
& \underline{93.29} & 79.82 & 81.03
& \textbf{70.22} & 76.24
& \underline{82.07} & 80.44
& 52.19 & 64.19 & 44.62
& 53.67 & 67.06 \\

\rowcolor{scrollStripeBg}
\hspace{0.9em}+D-HSM~\cite{dhsm}
& 1
& \underline{93.29} & 82.57
& \textbf{84.48} & 66.29
& 72.28 & 81.52 & 80.07
& 51.52 & 63.51 & 55.38
& 56.80 & 68.44 \\

\hspace{0.9em}+OASIS~\cite{oasis}
& 2
& 92.62 & \underline{84.40} & 78.45
& \underline{68.54} & 70.30
& 75.00 & 78.22 & 53.54
& \underline{65.54} & 52.15
& 57.08 & 67.65 \\

\rowcolor{scrollOursBg}
\hspace{0.9em}\textbf{+SCROLL (Ours)}
& 1
& 92.62 & \underline{84.40} & 82.76
& \underline{68.54} & \textbf{78.22}
& 80.43 & \underline{81.16}
& \textbf{57.91} & \textbf{71.62}
& \underline{70.97} & \textbf{66.83}
& \textbf{74.00} \\

\hline
\end{tabular}
\end{adjustbox}
\endgroup
\vspace{-1.5em}
\end{table*}

%% file: tables/streaming_comparison.tex
\begin{table*}[t]
\centering
\caption{\textbf{Results on StreamingBench real-time visual understanding.}
Accuracy (\%). Avg. is question-weighted across the RTVU questions.
$\dagger$ denotes baselines reproduced by us; other baseline values
are taken from published reports.}
\label{tab:02-streaming-main}

\begingroup
\fontsize{8.5}{10}\selectfont
\renewcommand{\arraystretch}{1.12}
\setlength{\tabcolsep}{2.75pt}

\begin{adjustbox}{max width=\textwidth}
\begin{tabular}{lc*{11}{r}}
\hline

\rowcolor{scrollHeaderBg}
& &
\multicolumn{11}{c}{\textbf{Real-Time Visual Understanding}} \\
\cline{3-13}
\rowcolor{scrollHeaderBg}
\multirow{-2}{*}{\textbf{Model}}
& \multirow{-2}{*}{\textbf{FPS}}
& \textbf{OP} & \textbf{CR} & \textbf{CS}
& \textbf{ATP} & \textbf{EU} & \textbf{TR}
& \textbf{PR} & \textbf{SU} & \textbf{ACP}
& \textbf{CT} & \textbf{Avg} \\
\hline

Human~\cite{zeng2025streamforest}
& --
& 89.47 & 92.00 & 93.60 & 91.47
& 95.65 & 92.52 & 88.00 & 88.75
& 89.74 & 91.30 & 91.46 \\

\hline
\rowcolor{scrollGroupBg}
\multicolumn{13}{l}{
\textit{\textbf{Proprietary Models}}} \\

Gemini 1.5 Pro~\cite{zeng2025streamforest}
& 1
& \textbf{79.02} & \textbf{80.47} & \underline{83.54}
& \textbf{79.67} & \textbf{80.00} & \textbf{84.74}
& \textbf{77.78} & \textbf{64.23} & \textbf{71.95}
& \underline{48.70} & \textbf{75.69} \\

\rowcolor{scrollStripeBg}
GPT-4o~\cite{zeng2025streamforest}
& --
& \underline{77.11} & \textbf{80.47} & \textbf{83.91}
& \underline{76.47} & \underline{70.19} & \underline{83.80}
& \underline{66.67} & \underline{62.19} & \underline{69.12}
& \textbf{49.22} & \underline{73.28} \\

\hline
\rowcolor{scrollGroupBg}
\multicolumn{13}{l}{
\textit{\textbf{Open-source Online MLLMs (Training-Based)}}} \\

AURA (8B)~\cite{aura2026}
& 2
& \textbf{87.50} & \underline{84.40} & \textbf{93.40}
& \textbf{89.50} & \underline{83.20} & \textbf{87.90}
& \textbf{86.10} & \textbf{76.40} & \textbf{82.40}
& 47.70 & \textbf{83.20} \\

\rowcolor{scrollStripeBg}
StreamScout-R (8B)~\cite{zhang2026streamscout}
& --
& -- & -- & -- & -- & -- & --
& -- & -- & -- & -- & \underline{82.80} \\

SelectStream (8B)~\cite{selectstream}
& 1
& \underline{86.30} & \textbf{87.40} & \underline{86.00}
& \underline{87.10} & \textbf{84.70} & \underline{84.60}
& \underline{85.80} & \underline{75.00} & \underline{78.20}
& \textbf{71.60} & 82.67 \\

\rowcolor{scrollStripeBg}
StreamTTT (4B)~\cite{chen2026streamttt}
& 2
& -- & -- & -- & -- & -- & --
& -- & -- & -- & -- & 80.48 \\

StreamForest (7B)~\cite{zeng2025streamforest}
& 1
& 83.11 & 82.81 & 82.65 & 84.26
& 77.50 & 78.19 & 76.85 & 69.11
& 75.64 & \underline{54.40} & 77.26 \\

\hline
\rowcolor{scrollGroupBg}
\multicolumn{13}{l}{
\textit{\textbf{Open-source Online MLLMs (Training-Free)}}} \\

SimpleStream (8B)~\cite{shen2026simplestream}
& 1
& -- & -- & -- & -- & -- & --
& -- & -- & -- & -- & 80.59 \\

\rowcolor{scrollStripeBg}
StreamScout (8B)~\cite{zhang2026streamscout}
& --
& -- & -- & -- & -- & -- & --
& -- & -- & -- & -- & 80.30 \\

HERMES (7B)~\cite{zhang2026hermes}
& 1
& -- & -- & -- & -- & -- & --
& -- & -- & -- & -- & 79.44 \\

\rowcolor{scrollStripeBg}
ObjectStream (7B)~\cite{dong2026objectstream}
& 1
& 74.10 & 80.20 & 88.60 & 82.60
& 77.10 & 85.00 & \underline{82.40}
& 73.60 & 73.20 & 44.20 & 76.80 \\

FluxMem (7B)~\cite{xie2026fluxmem}
& 1
& 80.20 & \underline{81.10} & 81.40
& 85.30 & 78.00 & 83.80 & 80.60
& 65.90 & 69.60 & \textbf{52.10} & 76.40 \\

\rowcolor{scrollStripeBg}
QueryStream (7B)~\cite{zhang2026querystream}
& 1
& 82.11 & \textbf{83.59} & 78.23
& 82.69 & 75.47 & 80.06 & 79.63
& 63.01 & 67.90 & 42.55 & 74.04 \\

\rowcolor{scrollGroupBg}
\multicolumn{13}{l}{
\textit{\textbf{Qwen2.5-VL-7B$^{\dagger}$}}~
\cite{bai2025qwen25vl}} \\

\hspace{0.9em}+Recent8
& 1
& 87.26 & 75.78 & \textbf{95.27}
& 90.35 & \underline{79.75}
& \underline{93.77} & 79.63
& \textbf{80.89} & 78.98
& 42.55 & 82.95 \\

\rowcolor{scrollOursBg}
\hspace{0.9em}+\textbf{SCROLL (Ours)}
& 1
& \textbf{88.62} & 77.34 & 94.64
& 90.35 & \textbf{81.65}
& \textbf{94.39} & 80.56
& \textbf{80.89} & 79.26
& 46.28 & \underline{83.71} \\

\rowcolor{scrollGroupBg}
\multicolumn{13}{l}{
\textit{\textbf{Qwen3-VL-8B$^{\dagger}$}}~
\cite{bai2025qwen3vl}} \\

\hspace{0.9em}+Recent8
& 1
& 86.72 & 74.22 & \textbf{95.27}
& \underline{92.28} & 79.11
& 91.59 & \textbf{86.11}
& \underline{79.67} & \underline{84.66}
& 42.02 & 83.63 \\

\rowcolor{scrollOursBg}
\hspace{0.9em}+\textbf{SCROLL (Ours)}
& 1
& \underline{88.08} & 75.00
& \underline{94.95} & \textbf{92.60}
& \underline{79.75} & 92.52
& \textbf{86.11} & \underline{79.67}
& \textbf{84.94} & \underline{47.34}
& \textbf{84.47} \\

\hline
\end{tabular}
\end{adjustbox}
\endgroup
\vspace{-1.5em}
\end{table*}

%% file: sec/4_experiment.tex
\section{Experiments}
\label{sec:experiments}

\noindent \textbf{Implementation Details.}
Our experiments use Qwen3-VL-8B-Instruct~\cite{bai2025qwen3vl} and
Qwen2.5-VL-7B-Instruct~\cite{bai2025qwen25vl} as the VLM backbones.
For SCROLL-Index, we sample video frames at 1 FPS and use
CLIP ViT-B/32~\cite{radford2021clip} to extract 512-dimensional normalized
embeddings, each linked to its source frame and timestamp.
For SCROLL-Explorer, we initialize the visual context with
eight recent frames unless otherwise specified.
The ASK loop allows at most six rounds and retains up to
eight historical frames.
The same VLM instance performs evidence assessment,
search planning, selective retention, and answer generation.
Search queries are encoded by the CLIP text encoder,
and each search retrieves up to two unseen historical
frames using cosine similarity within the requested
temporal range.
We use greedy decoding for VLM generation, with BF16 precision
and FlashAttention-2~\cite{dao2024flashattention2} for inference.
On OVO-Bench, the 8B model returns the terminal ASK answer;
the 7B model and both StreamingBench settings additionally
perform a final QA over the selected frames after exploration.
All experiments run on NVIDIA RTX A5000 GPUs,
with each question processed on a single GPU.

\subsection{Benchmarks}

We evaluate SCROLL on OVO-Bench~\cite{niu2025ovobench} and
StreamingBench~\cite{lin2026streamingbench}. On both benchmarks, under a strict online setting, the model answers each multiple-choice question using only frames
available up to the query time.

\noindent \textbf{OVO-Bench.}
We evaluate the Real-Time Visual Perception (RT) and Backward
Tracing (BW) subsets, which together contain 1,468 questions.
RT comprises 837 questions across six subtasks that assess the
model's understanding of the current scene. BW comprises 631
questions across three subtasks: Episodic Memory (EPM), Action
Sequence Identification (ASI), and Hallucination Detection (HLD).
These test whether the model can recall earlier events, recover
the order of past actions, and recognize when the video does not
support a question's premise.

\noindent \textbf{StreamingBench.}
We evaluate the Real-Time Visual Understanding track of StreamingBench, which covers
ten tasks ranging from object and action perception to spatial
understanding, causal reasoning, and counting. We report accuracy
for each task and compute the overall score as the fraction of
questions answered correctly.

\subsection{Main Results}

\noindent \textbf{Results on OVO-Bench.}
Table~\ref{tab:01-ovo-main} demonstrates that SCROLL substantially improves historical understanding while preserving strong real-time perception.
With Qwen3-VL-8B, SCROLL achieves a 13.16-point gain in backward tracing over Recent8, while maintaining a real-time perception accuracy of 81.16\%.
This yields the highest overall accuracy of 74.00\% among the compared methods, surpassing both training-free and training-based baselines.
Similar improvements with Qwen2.5-VL-7B further support the effectiveness of combining lightweight visual memory with adaptive evidence exploration.

\noindent \textbf{Results on StreamingBench.}
Table~\ref{tab:02-streaming-main} further demonstrates the effectiveness of SCROLL across diverse real-time visual understanding tasks.
SCROLL achieves the highest average accuracy of 84.47\% among the compared methods, with consistent improvements over recent-window baselines across both backbones.
Together, these results highlight that on-demand access to historical evidence can complement strong perception of the ongoing video.
\input{tables/ASK_ablation}

\input{tables/answer_first_last}
\input{tables/retention_strategies}

\begin{figure}[t]
    \centering
    \includegraphics[width=0.7\linewidth]{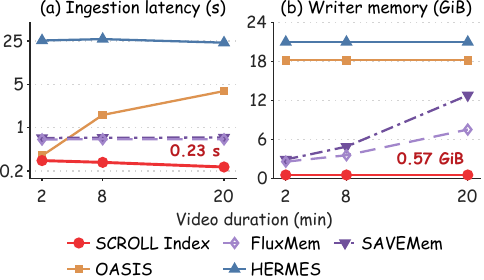}
    \caption{\textbf{Efficiency of streaming memory construction.}
    Mean frame ingestion latency (a) and peak writer GPU memory (b) on
    2-, 8-, and 20-minute video prefixes.}
    \label{fig:ingestion-cost}
    \vspace{-1.75em}
\end{figure}
\begin{figure*}[t]
    \centering
    \includegraphics[width=\linewidth]{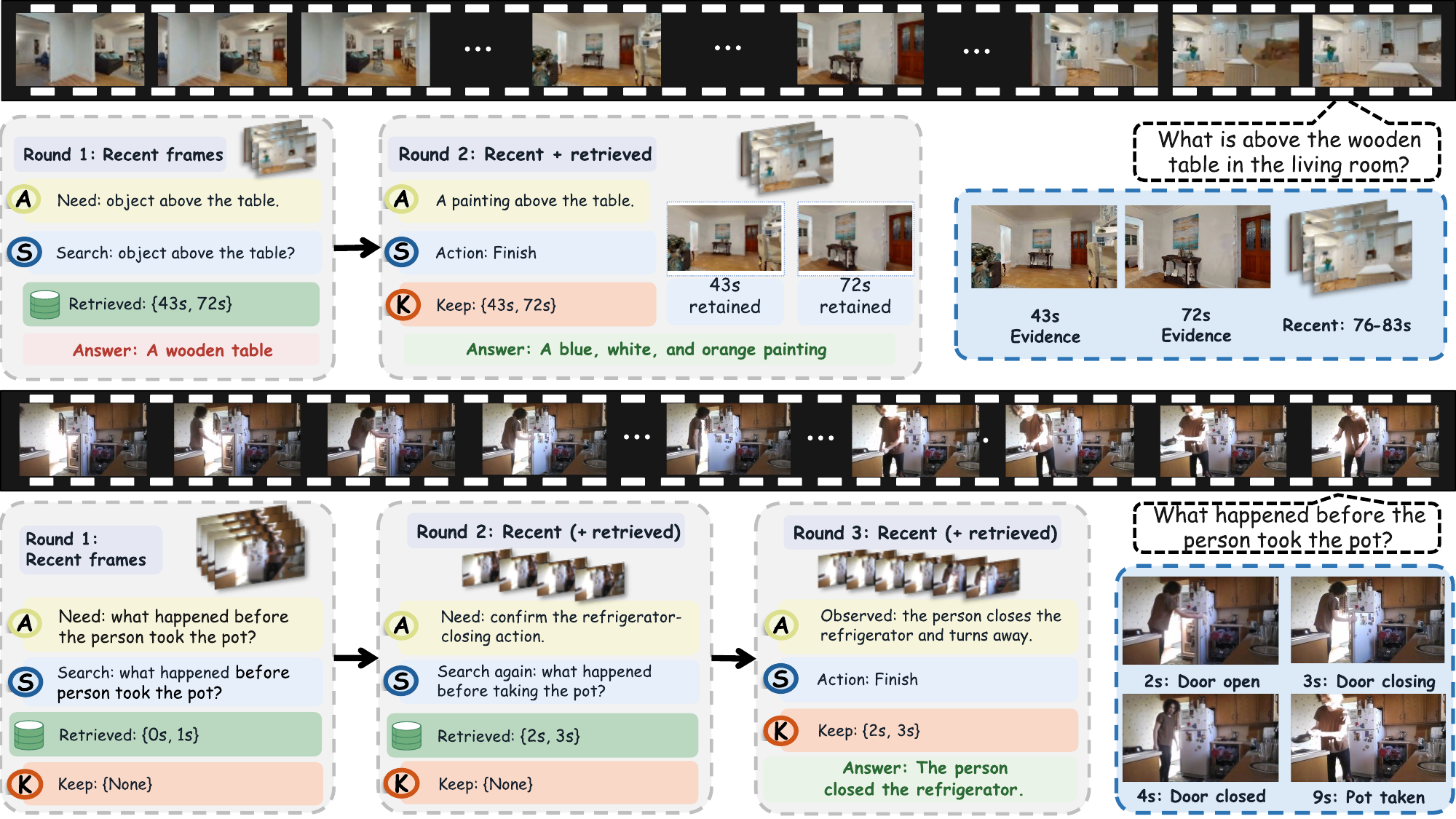}
\caption{\textbf{Question-driven historical evidence acquisition with SCROLL.}
Top: retrieving earlier living-room views identifies the painting above the table and corrects the initial answer.
Bottom: successive searches recover the refrigerator-closing action before pot-taking.
These illustrate how ASK identifies missing information and acquires historical visual evidence for spatial and temporal reasoning.}
    \label{fig:ask-cases}
    \vspace{-1.5em}
\end{figure*}

\subsection{Ablation Studies}

\noindent\textbf{Contributions of the ASK loop.}
Table~\ref{tab:ask-components} demonstrates the
effectiveness of adaptive evidence exploration.
Starting from Recent8, a single-round ASK variant
without historical retrieval achieves 71.75\%
Overall accuracy, while the full ASK process
improves this to 74.00\%.
Compared with random historical retrieval,
evidence-guided Search achieves comparable
accuracy with 13.1\% fewer VLM calls and
13.6\% fewer cumulative input tokens.
Removing Assess reduces Overall accuracy
by 2.98 points despite 38\% more Reader calls,
highlighting its role in regulating exploration.
Together, these results demonstrate the
benefits of adaptive evidence acquisition
and context management.

\noindent\textbf{Answer refinement during exploration.}
Table~\ref{tab:answer-refinement} examines how
answers evolve within the same SCROLL trajectories.
Initial ASK predictions already improve over
the official Recent8 baseline, particularly
on hallucination detection.
Subsequent exploration further improves
episodic memory and action sequence identification,
highlighting its value for recovering historical
events and temporal relationships.
These results suggest that the benefits of
evidence exploration vary across tasks,
underscoring the importance of selective
historical context utilization.

\noindent\textbf{Accuracy--cost trade-off of visual retention.}
As shown in Table~\ref{tab:retention-cost}, selective retention achieves a favorable accuracy--cost trade-off by maintaining useful historical context with fewer input tokens. It achieves the highest overall accuracy while reducing cumulative input tokens across all reasoning rounds by 14.7\% compared with keep-all, and outperforms FIFO under a matched history capacity.

\noindent \textbf{Streaming memory efficiency.}
Figure~\ref{fig:ingestion-cost} shows that SCROLL
maintains low writing latency and stable GPU memory usage
as video duration increases.
On 20-minute videos, SCROLL achieves nearly 3$\times$
lower per frame writing latency and 13--22$\times$ lower peak GPU
memory usage than FluxMem and SAVEMem.

We further evaluate concurrent indexing and question
answering on the same RTX A5000.
Across paired OVO-Bench questions, continuous 1-FPS
indexing adds only 0.19\,s to the mean end-to-end QA latency, with identical predictions and
VLM call traces.


\subsection{Qualitative Analysis}

Figure~\ref{fig:ask-cases} illustrates SCROLL's
question-driven historical evidence acquisition.
In the upper example, the Recent frames (76--83\,s)
do not reveal the object above the wooden table.
ASK identifies this missing information, retrieves
earlier living-room views at 43\,s and 72\,s,
and corrects its initial answer to a blue, white,
and orange painting.

In the lower example, the question asks which
action preceded pot-taking at 9\,s.
The first retrieval (0--1\,s) is insufficient
to resolve the sequence, prompting another search.
The retrieved frames at 2--3\,s show the
refrigerator door closing and allow the Reader
to establish the temporal order in the third round.
Once an answer is supported, Keep selects relevant
historical frames, leaving uninformative retrieved
frames outside the retained context.


%% file: tables/ASK_ablation.tex
\begin{table}[t]
\centering
\caption{\textbf{Ablation of ASK components on OVO-Bench.}
Results use Qwen3-VL-8B. A/S/K denote Assess,
Search, and Keep. Agent variants start from Recent8
and allow up to six rounds; random retrieval samples
historical frames. The arrow indicates the gain
over Recent8 (pp).}
\label{tab:ask-components}

\begingroup
\small
\renewcommand{\arraystretch}{1.3}
\setlength{\tabcolsep}{4.5pt}

\begin{adjustbox}{max width=\linewidth}
\begin{tabular}{lcccrrcr}
\hline
\rowcolor{scrollHeaderBg}
\textbf{Variant}
& \textbf{A} & \textbf{S} & \textbf{K}
& \textbf{RT} & \textbf{BW}
& \textbf{Overall} & \textbf{VLM calls} \\
\hline

Recent8
& -- & -- & --
& 80.44 & 53.67 & 67.06 & 1 \\

\rowcolor{scrollStripeBg}
Recent8+CLIP Top-8
& -- & -- & --
& 79.13 & 52.46 & 65.80 & 1 \\

\hline

Recent8 + ASK (1 round)
& \yes & -- & --
& 79.68 & 63.81 & 71.75 & 1.015 \\

\rowcolor{scrollStripeBg}
w/o Assess
& \no & \yes & \yes
& 79.81 & 62.22 & 71.02 & 4.456 \\

Random retrieval
& \yes & \no & \yes
& 80.77 & 65.35 & 73.06 & 3.708 \\

\rowcolor{scrollStripeBg}
w/o Keep
& \yes & \yes & \no
& 81.02 & 66.07 & 73.54 & 3.213 \\

\rowcolor{scrollOursBg}
\textbf{Full SCROLL}
& \yes & \yes & \yes
& \textbf{81.16}
& \textbf{66.83}
& \textbf{74.00}\gain{6.94}
& 3.221 \\

\hline
\end{tabular}
\end{adjustbox}
\endgroup
\vspace{-1em}
\end{table}

%% file: tables/answer_first_last.tex
\begin{table}[t]
\centering
\caption{\textbf{Answer refinement through SCROLL exploration.}
Initial and final answers are evaluated within the same
Qwen3-VL-8B trajectories on OVO-Bench backward tracing,
up to six ASK rounds.
Arrows denote gains over initial answers (pp).}
\label{tab:answer-refinement}

\begingroup
\small
\renewcommand{\arraystretch}{1.3}
\setlength{\tabcolsep}{8pt}

\newcommand{\finalentry}[2]{%
    \makebox[3.0em][c]{#1}%
    \makebox[3.5em][c]{\hspace{0.2em}#2}%
}

\begin{adjustbox}{max width=\linewidth}
\begin{tabular}{lrcc}
\hline

\rowcolor{scrollHeaderBg}
& &
\multicolumn{2}{c}{\textbf{SCROLL}} \\
\cline{3-4}

\rowcolor{scrollHeaderBg}
\multirow{-2}{*}{\textbf{Subtask}}
& \multirow{-2}{*}{\textbf{Recent8}}
& \textbf{Initial answer}
& \textbf{Final answer} \\

\hline
EPM
& 52.19 & 54.21
& \finalentry{\textbf{57.91}}{} \\

\rowcolor{scrollStripeBg}
ASI
& 64.19 & 64.86
& \finalentry{\textbf{71.62}}{} \\

EPM/ASI Avg.
& 58.19 & 59.54
& \finalentry{\textbf{64.77}}{\gain{5.23}} \\

\hline
HLD
& 44.62 & \textbf{75.81}
& \finalentry{70.97}{} \\

\rowcolor{scrollOursBg}
\textbf{BW Avg.}
& 53.67 & 64.96
& \finalentry{\textbf{66.83}}{\gain{1.87}} \\

\hline
\end{tabular}
\end{adjustbox}
\endgroup
\vspace{-0.9em}
\end{table}

%% file: tables/retention_strategies.tex
\begin{table}[t]
\centering
\caption{\textbf{Accuracy--cost trade-off of visual retention.}
All variants use Qwen3-VL-8B, Recent8, and at most six
ASK rounds on OVO-Bench.
Input tokens are counted per-query cumulatively.
Keep-all imposes no limit on historical context capacity.}
\label{tab:retention-cost}

\begingroup
\small
\renewcommand{\arraystretch}{1.3}
\setlength{\tabcolsep}{6pt}

\begin{adjustbox}{max width=\linewidth}
\begin{tabular}{lcrr}
\hline

\rowcolor{scrollHeaderBg}
\textbf{Retention policy}
& \textbf{\shortstack{History\\capacity}}
& \textbf{Overall}
& \textbf{\shortstack{Input tokens\\per query}} \\

\hline
Keep-all
& Unlimited & 73.89 & 23,239 \\

\rowcolor{scrollStripeBg}
FIFO
& 4 & 72.23 & 21,088 \\

Selective
& 4 & 72.76 & 19,420 \\

\rowcolor{scrollOursBg}
\textbf{Selective (Full)}
& 8 & \textbf{74.00} & 19,826 \\

\hline
\end{tabular}
\end{adjustbox}
\endgroup
\vspace{-0.75em}
\end{table}

%% file: sec/5_conclusion.tex
\section{Conclusion}
\label{sec:conclusion}

In this work, we present SCROLL, a training-free framework that decouples streaming-time memory construction from query-time evidence exploration. SCROLL-Index maintains a lightweight, source-linked visual index, while SCROLL-Explorer uses an iterative Assess–Search–Keep (ASK) loop to retrieve and retain relevant historical evidence on demand. Experiments on OVO-Bench and StreamingBench demonstrate improved historical understanding, strong real-time perception, and substantially reduced memory construction overhead. More broadly, SCROLL highlights the potential of agentic retrieval for streaming video understanding and motivates co-adapting memory designs and adaptive evidence exploration across long, open-ended video streams.

%% file: sec/6_LLM_usage.tex
\section*{LLM Usage Statement}
OpenAI ChatGPT assisted with manuscript drafting,
source-code inspection, implementation support, \LaTeX \quad file preparation,
experimental result organization, and plotting code and figure layout.
Reported numerical results are derived from recorded experiments. The models used within SCROLL and the evaluated baselines,
together with their roles and inference settings, are described
in the method, experiments, and supplementary material.
The authors have inspected and verified relevant data, and retain full responsibility for the experimental protocols, code, figures, and reported results.

%% file: sec/X_suppl.tex

\clearpage
\setcounter{page}{1}
\maketitlesupplementary
\setcounter{section}{0}
\renewcommand{\thesection}{\Alph{section}}
\setcounter{table}{0}
\renewcommand{\thetable}{S\arabic{table}}
\ifdefined\theHsection
  \renewcommand{\theHsection}{supp.\arabic{section}}
\fi
\ifdefined\theHtable
  \renewcommand{\theHtable}{supp.\arabic{table}}
\fi

\section{Additional Experimental Analysis}
\label{sec:sup-results}

\noindent\textbf{Experimental and Ablation Protocols.}
Unless otherwise specified, experiments in this section
use Qwen3-VL-8B on OVO-Bench. The default SCROLL
configuration starts with eight recent frames (Recent8),
performs no historical pre-retrieval, and allows at most
six Assess--Search--Keep (ASK) rounds. Reported VLM calls
and input tokens are cumulative per-query averages,
including formatting retries.

We distinguish three answering settings.
The official Recent8 baseline uses the benchmark's
single-call QA protocol. Recent8+ASK (1 round) instead
performs a single structured ASK Reader invocation
on the same recent frames without historical retrieval,
using the final-round instruction described in
Sec.~\ref{sec:sup-ask-prompt}.
In contrast, the initial answers in
Table~\ref{tab:answer-refinement} come from the first
Reader invocation of full SCROLL trajectories, before
historical retrieval and without the final-round
instruction.

For the component ablations in
Table~\ref{tab:ask-components}, w/o Assess removes
explicit evidence-sufficiency assessment, while
w/o Keep disables selective retention of historical
frames across rounds. Random retrieval replaces
evidence-guided frame selection with random historical
sampling, with results averaged over four seeds.
The S indicator denotes evidence-guided Search but not the presence of historical retrieval itself.
For Qwen3-VL-8B on OVO-Bench, the terminal ASK answer
is returned directly.
Other backbone-specific answering protocols are detailed
in Sec.~\ref{sec:sup-readout}.

\subsection{Query-Time Visual Budget Analysis}
\label{sec:sup-budgets}

\noindent\textbf{On-demand visual context allocation.}
Table~\ref{tab:sup-budget} examines the budget allocated to
historical frames before the first ASK decision, the Recent-frame
window, and the maximum number of exploration rounds.
Starting from Recent frames alone achieves the highest Overall
accuracy among the initial-history settings (74.00\%) while using
14.1\% fewer input tokens than pre-retrieving eight historical
frames. The default Recent8 window achieves the strongest Overall
accuracy among the tested Recent-frame budgets. Allowing six ASK
rounds yields the highest Overall accuracy while requiring an average
of 3.221 VLM calls per question. These results support allocating
historical evidence on demand while preserving recent visual context.

\begin{table}[t]
\centering
\caption{\textbf{Query-time visual budget analysis on OVO-Bench.}
The three panels vary (a) initially retrieved historical frames,
(b) Recent frames, and (c) maximum ASK rounds. Input tokens are
cumulative per-question means (in thousands). Blue rows denote
the default configuration.}
\label{tab:sup-budget}

\begingroup
\small
\renewcommand{\arraystretch}{1.13}
\setlength{\tabcolsep}{4pt}

\begin{adjustbox}{max width=\linewidth}
\begin{tabular}{lrrr}
\hline

\rowcolor{scrollHeaderBg}
\textbf{Setting}
& \textbf{Overall}
& \textbf{VLM calls}
& \textbf{Input tokens ($10^3$)} \\

\hline
\rowcolor{scrollGroupBg}
\multicolumn{4}{l}{
\textit{(a) Initial historical frames}} \\

\rowcolor{scrollOursBg}
\textbf{0 (default)}
& \textbf{74.00} & \textbf{3.221}
& \textbf{19.83} \\

2 & 72.81 & 3.401 & 22.22 \\

\rowcolor{scrollStripeBg}
4 & 72.81 & 3.226 & 22.05 \\

8 & 73.74 & 3.182 & 23.08 \\

\hline
\rowcolor{scrollGroupBg}
\multicolumn{4}{l}{
\textit{(b) Recent frames}} \\

0 & 59.83 & 4.694 & 13.48 \\

\rowcolor{scrollStripeBg}
2 & 71.78 & 3.645 & 11.57 \\

4 & 72.11 & 3.679 & 15.46 \\

\rowcolor{scrollOursBg}
\textbf{8 (default)}
& \textbf{74.00} & \textbf{3.221}
& \textbf{19.83} \\

\hline
\rowcolor{scrollGroupBg}
\multicolumn{4}{l}{
\textit{(c) Maximum ASK rounds}} \\

1 & 71.75 & 1.015 & 5.24 \\

\rowcolor{scrollStripeBg}
2 & 72.74 & 1.617 & 8.81 \\

3 & 72.49 & 2.063 & 11.71 \\

\rowcolor{scrollStripeBg}
4 & 72.97 & 2.470 & 14.49 \\

5 & 72.82 & 2.847 & 17.16 \\

\rowcolor{scrollOursBg}
\textbf{6 (default)}
& \textbf{74.00} & \textbf{3.221}
& \textbf{19.83} \\

\hline
\end{tabular}
\end{adjustbox}
\endgroup
\end{table}

\subsection{Uniform Historical Sampling}
\label{sec:sup-uniform}

\noindent\textbf{Uniform sampling versus adaptive exploration.}
Table~\ref{tab:sup-uniform} examines whether directly
adding historical frames improves visual understanding.
Under the same one-round ASK protocol, uniform sampling
increases backward-tracing accuracy but does not improve
Overall accuracy, despite nearly doubling input tokens.
SCROLL achieves higher RT, BW, and Overall accuracy
through adaptive evidence exploration.

\begin{table}[t]
\centering
\caption{\textbf{Uniform historical sampling on OVO-Bench.}
All variants use Qwen3-VL-8B and Recent8.
Uniform10 adds up to ten uniformly sampled historical frames
before the first ASK decision.
The one-round variants share the same answering protocol,
while full SCROLL allows up to six ASK rounds.
VLM calls and input tokens are cumulative per-query means,
including formatting retries.}
\label{tab:sup-uniform}

\begingroup
\small
\renewcommand{\arraystretch}{1.15}
\setlength{\tabcolsep}{3pt}

\begin{adjustbox}{max width=\linewidth}
\begin{tabular}{lrrrrr}
\hline

\rowcolor{scrollHeaderBg}
\textbf{Setting}
& \textbf{RT}
& \textbf{BW}
& \textbf{Overall}
& \textbf{\shortstack{VLM\\calls}}
& \textbf{\shortstack{Input\\tokens (K)}} \\

\hline
SCROLL (1 round)
& 79.68 & 63.81 & 71.75 & 1.015 & 5.24 \\

\rowcolor{scrollStripeBg}
\hspace{0.9em}+Uniform10
& 78.39 & 64.52 & 71.46 & 1.012 & 10.37 \\

\rowcolor{scrollOursBg}
\textbf{SCROLL (Full)}
& \textbf{81.16}
& \textbf{66.83}
& \textbf{74.00}
& 3.221
& 19.83 \\

\hline
\end{tabular}
\end{adjustbox}
\endgroup
\end{table}

\subsection{Paired Accuracy Gains on StreamingBench}
\label{sec:sup-paired}

\noindent\textbf{Consistent improvements across VLM backbones.}
Table~\ref{tab:sup-paired} complements the main-paper StreamingBench
results with paired accuracy intervals. SCROLL improves over Recent8
on both Qwen2.5-VL-7B and Qwen3-VL-8B, with positive 95\% confidence
intervals under video-cluster resampling.

\begin{table}[t]
\centering
\caption{\textbf{Paired accuracy gains on StreamingBench.}
Accuracy (\%) on the Real-Time Visual Understanding track.
Confidence intervals use 10,000 paired video-cluster bootstrap
resamples over the same questions. Gains are in pp.}
\label{tab:sup-paired}

\begingroup
\small
\renewcommand{\arraystretch}{1.14}
\setlength{\tabcolsep}{4pt}

\begin{adjustbox}{max width=\linewidth}
\begin{tabular}{lrrrr}
\hline

\rowcolor{scrollHeaderBg}
\textbf{VLM backbone}
& \textbf{Recent8}
& \textbf{SCROLL}
& \textbf{Gain}
& \textbf{95\% CI} \\

\hline
Qwen2.5-VL-7B
& 82.95
& \textbf{83.71}
& \textcolor{scrollGain}{\textbf{+0.76}}
& [0.20, 1.36] \\

\rowcolor{scrollStripeBg}
Qwen3-VL-8B
& 83.63
& \textbf{84.47}
& \textcolor{scrollGain}{\textbf{+0.84}}
& [0.36, 1.36] \\

\hline
\end{tabular}
\end{adjustbox}
\endgroup
\end{table}

\section{Implementation Details}
\label{sec:sup-implementation}

\subsection{Evidence Retrieval and Context Management}
\label{sec:sup-retrieval}

SCROLL-Index maintains CLIP embeddings linked to source frames and
timestamps. During Evidence-Guided Historical Search, SCROLL-Explorer
considers up to 32 similarity-ranked eligible historical frame IDs
and returns at most two previously unseen frames per search.
It prefers candidates from distinct four-frame temporal bins; an
optional temporal anchor and direction restrict the eligible period.
Searches exclude already displayed historical frames and obey the
fixed query-time cutoff. Newly selected Retained frames are prioritized
before earlier retained evidence, subject to the eight-frame historical
capacity. Evicting a historical frame from the visual context does not delete
its source observation from SCROLL-Index. The evaluated configuration does not use
cropping or a separate similarity acceptance threshold for ASK search.

\subsection{Answer Readout and Evaluation Settings}
\label{sec:sup-readout}

On OVO-Bench, Qwen3-VL-8B returns the answer from the terminal ASK
response, while Qwen2.5-VL-7B uses an additional final QA over the
terminal visual context, the original question, and the final
unresolved information need. On StreamingBench,
both VLM backbones apply the official multiple-choice QA prompt after
ASK exploration to Recent and terminal Retained frames; newly Retrieved
frames enter this readout only if retained. The final QA performs no
additional retrieval. A formatting retry, if required, reuses the same
visual inputs. The OVO-Bench Overall score averages the Real-Time Visual Perception
(RT) and Backward Tracing (BW) subtask means; the StreamingBench
Avg. is question-weighted.
All observations are restricted to the causal video prefix available
at the question time. Memory-writing measurements in the main paper
separate full-prefix construction from clock-paced ingestion: the
former starts with decoded frames, while the latter includes frame
arrival, queueing, and decoding. Comparisons use the evaluated native
settings of each baseline.

\section{Core Prompt Templates}
\label{sec:sup-prompts}

The prompts below reproduce the executed core instruction text; braces
and angle brackets denote runtime-filled fields. The original question
and answer options are inserted in the named fields. For each ASK
call, images precede the prompt in frame-ID order with source-time
labels. The recorded search history supplies prior tool requests and
observations, and the query-time cutoff remains fixed throughout ASK.
Benchmark-provided one-call QA prompts and formatting-only retries are
not repeated here. The additional final QA readouts described in
Sec.~\ref{sec:sup-readout} are applied after exploration and do not
change the executed ASK decisions.

\lstdefinestyle{scrollprompt}{
  basicstyle=\ttfamily\fontsize{6.5}{7.3}\selectfont,
  columns=fullflexible,
  breaklines=true,
  breakatwhitespace=false,
  keepspaces=true,
  showstringspaces=false,
  frame=single,
  framerule=0.35pt,
  rulecolor=\color{black!38},
  backgroundcolor=\color{black!3},
  framesep=4pt,
  xleftmargin=0pt,
  xrightmargin=0pt,
  aboveskip=2pt,
  belowskip=2pt
}

\subsection{Shared ASK Prompt}
\label{sec:sup-ask-prompt}

\begin{lstlisting}[style=scrollprompt,title={Prompt 1: Shared ASK instruction}]
Inspect the actual images and decide whether another tool is needed to answer this question. The answer choices are provided to clarify what evidence the question requires.
Question: {question}
Fixed task interpretation (fallible; original question takes precedence): {question}
Preserve the original reference action and the requested relation throughout retrieval.
Answer choices:
{answer_choices}
Question time: {question_time:.3f} seconds. Never request future observations.
Each image has a frame ID and actual source time. Recent means the recent view, not all past events.

First identify the question's unknown fact and any reference event. Preserve which object belongs to which action. An unknown object or action remains unknown; do not bind it to the reference object just because that object is named in the question.
Record at most TWO short observations with supporting displayed frame IDs. These are fallible visual observations, not assumed facts. A relevant object or background does not establish an action, its receiving object, identity, or time. Do not turn a question premise into an observed event.
Before finishing, state the answering fact itself in observations and cite the displayed frames that support it. Check that this fact answers the ENTIRE original question about the requested entity, relationship and time. Naming a related object, scene or reference action alone is insufficient. Clear need only when the requested fact is supported; otherwise keep the specific unanswered fact in need and choose a tool that could resolve it. Once the answer is supported, finish without searching for unasked facts. If no productive tool remains, finish with need still nonempty.
For search, anchor is a TIME reference: use before/after only after the reference event is supported in displayed images; otherwise use anchor=-1 and direction=both. Keep the reference event and the unknown answer event separate. A current view does not prove an earlier event, and an old observation does not prove the current state.

Tools:
- search: Search the written historical key index again, covering frame IDs 0 through {history_cutoff_minus_one}; each query can retrieve frames outside the initial candidate pool. Return at most TWO unseen historical frames for one concise visual query. anchor=-1 searches all available history; otherwise before/after restricts time relative to a displayed anchor. Scores identify candidates, not verified facts.
- finish: stop when required facts have supporting images, or when no productive search remains. In the latter case keep need nonempty. No retrieved match does not prove that an event never occurred.

Previously attempted tools and observations:
{history}
Allowed displayed integer frame IDs: {displayed}.
Historical frame IDs eligible for keep: {historical}. Recent is retained automatically.
Inspection {step} of at most {maximum_checks}. {last_step}
Output exactly one compact JSON object, with ALL NINE fields, including your current answer:
{"keep":[integer_IDs],"observations":[{"frames":[integer_ID],"fact":"at most 12 words describing visible evidence"}],"need":"one specific unresolved fact, at most 20 words, or empty if supported","action":"search|finish","query":"one visual search phrase, at most 12 words, or empty","anchor":-1,"region":0,"direction":"before|after|both","answer":"one offered option letter"}
At most two observations. Keep at most six useful displayed historical frames, including complementary before/after states. Empty observations and keep lists are allowed. Use integer frame IDs, not timestamps. Search requires a nonempty need. Do not output tool-result fields.
Mandatory fields: keep, observations, need, action, query, anchor, region, direction, answer. ALWAYS output keep, including keep=[] when no historical image should be retained. Previously retained history persists; keep places this round's useful evidence first within the eight-frame retained-history limit. Use a single compact JSON line.

Previously retained historical frame IDs: {retained}. Keep useful reference and answer states.
Write answer LAST, after the evidence and search decision: your current best offered option letter using only the images displayed in this call. Always provide one answer, including when further search is needed. If action is finish or the execution budget ends, this answer is returned directly; there is no separate final answering call. Providing a best current answer does not establish evidence or justify clearing need. Keep unresolved information explicit.
\end{lstlisting}

\noindent\textbf{Final-round instruction.}
On the final permitted ASK round, the \texttt{\{last\_step\}}
field is replaced by the following executed instruction:
\emph{``No tool calls remain. Review the existing images, retain useful
historical evidence, and preserve any unresolved need. Set action to
finish; exhausting the budget does not make evidence sufficient or
justify clearing useful history.''}

\subsection{Backbone-Specific Instructions}
\label{sec:sup-backbone-prompt}

Qwen2.5-VL-7B appends the following suffix to the shared ASK prompt.

\begin{lstlisting}[style=scrollprompt,title={Prompt 2: Qwen2.5-VL concise suffix}]
Answer from the actual displayed images and their times, matching the objects, actions, and relations asked about. The original question and displayed images take precedence over fallible notes. Keep all nine required JSON fields. Write at most two short observation sentences, each with its displayed frame IDs. Use one short, specific sentence for any unresolved need and a concise visual phrase for the retrieval query; leave fields empty only as the original protocol allows. Do not copy schema placeholders.
\end{lstlisting}

\subsection{Final Answer Readout}
\label{sec:sup-final-prompt}

For OVO-Bench with Qwen2.5-VL-7B, the following prompt reads the final
answer from the terminal ASK visual context. The \texttt{TERMINAL\_NEED}
field is populated by the final unresolved information need, or
\texttt{none reported} when empty.

\begin{lstlisting}[style=scrollprompt,title={Prompt 3: OVO-Bench 7B final QA}]
Answer the question using the video frames and their actual times. Match the requested object, action and temporal relationship. The interpretation and missing-evidence note below are fallible; the original question and images take precedence. For earlier events, the recent view alone does not establish what happened before. A similar object is not proof of an action; a missing match is not proof an event never occurred. Question premises are not observed facts. Return only the option letter.
Question: <QUESTION>
Options:
<OPTIONS_ONE_PER_LINE>
Task interpretation: <QUESTION>
Still unresolved: <TERMINAL_NEED>
\end{lstlisting}

The OVO-Bench Qwen3-VL-8B configuration reads the terminal ASK answer
directly. StreamingBench uses its official QA prompt on the selected
frames after ASK terminates.